%% file: acl_latex.tex
\pdfoutput=1

\documentclass[11pt]{article}

\usepackage[final]{acl}

\usepackage{times}
\usepackage{latexsym}
\usepackage{setspace} 

\usepackage[T1]{fontenc}

\usepackage[utf8]{inputenc}

\usepackage{microtype}

\usepackage{inconsolata}

\usepackage{graphicx}

\usepackage{multirow}
\usepackage[normalem]{ulem}
\useunder{\uline}{\ul}{}
\usepackage{fontawesome5}
\usepackage{booktabs}
\usepackage{graphicx}
\usepackage{color}
\usepackage{xcolor}
\usepackage{xspace}
\usepackage{subfigure}
\usepackage{todonotes}
\usepackage{enumitem}
\usepackage[most]{tcolorbox}
\usepackage{array}
\usepackage{cleveref}

\definecolor{bggray}{rgb}{0.95, 0.95, 0.95}
\usepackage[%
    framemethod=tikz,
    skipbelow=\topskip,
    skipabove=\topskip
]{mdframed}
\mdfsetup{%
    leftmargin=0pt,
    rightmargin=0pt,
    backgroundcolor=bggray,
    middlelinecolor=black,
    roundcorner=3
}
\newtcolorbox[list inside=prompt,auto counter,number within=section]{prompt}[1][]{
    colbacktitle=black!60,
    fonttitle=\small,
    coltitle=white,
    fontupper=\footnotesize,
    boxsep=4pt,
    left=0pt,
    top=0pt,
    bottom=0pt,
    boxrule=1pt,
    #1,
}

\title{Language Models Predict Empathy Gaps \\ Between Social In-groups and Out-groups}

\author{
    Yu Hou \quad
    Hal {Daum\'e III} \quad
    Rachel Rudinger \\
    University of Maryland \\
    \texttt{\{houyu,hal3,rudinger\}@umd.edu}
}

\begin{document}
\maketitle
\begin{abstract}
\input{pages/00_abs}
\end{abstract}

\input{pages/01_intro}
\input{pages/02_background}
\input{pages/03_exp-setup}
\input{pages/04_result}
\input{pages/05_analysis}
\input{pages/06_discussion}

\input{pages/11_limitation}
\input{pages/12_ethical}
\input{pages/13_ack}

\bibliography{custom}

\clearpage
\appendix
\input{appendix/00_prompt-variations}
\input{appendix/01_model-details}
\input{appendix/02_result-details}

\end{document}

%% file: pages/00_abs.tex
Studies of human psychology have demonstrated that people are more motivated to extend empathy to in-group members than out-group members~\cite{Cikara2011-us-and-them}.
In this study, we investigate how this aspect of intergroup relations in humans is replicated by LLMs in an emotion intensity prediction task.
In this task, the LLM is given a short description of an experience a person had that caused them to feel a particular emotion; the LLM is then prompted to predict the intensity of the emotion the person experienced on a numerical scale.
By manipulating the group identities assigned to the LLM's persona (the ``perceiver'') and the person in the narrative (the ``experiencer''), we measure how predicted emotion intensities differ between in-group and out-group settings.
We observe that LLMs assign higher emotion intensity scores to in-group members than out-group members.
This pattern holds across all three types of social groupings we tested: race/ethnicity, nationality, and religion.
We perform an in-depth analysis on Llama-3.1-8B, the model which exhibited strongest intergroup bias among those tested.\footnote{Code and data can be found at \url{https://github.com/houyu0930/intergroup-empathy-bias.}}

%% file: pages/01_intro.tex
\section{Introduction}

\expandafter\def\expandafter\quote\expandafter{\quote\small\setstretch{1}}
\begin{quote}
    {
    ``\textit{People are often motivated to increase others' positive experiences and to alleviate others' suffering ... When the target is an outgroup member, however, people may have powerful motivations not to care about or help that ``other''.}'' \\\phantom{------------------------}---- \citet{Cikara2011-us-and-them}
    }
\end{quote}

\input{figures/00_setup}

\noindent
As language technologies play an increasingly important role in interpersonal communication in society, research has shown that their use can impact social relationships~\cite{Hohenstein2023-communication}.
This could potentially occur when communication partners perceive one another differently through their use of suggestions from assistant tools (e.g. ChatGPT).
This impact on social relationships can be exacerbated because people are cognitive misers~\cite{fiske1991social, STANOVICH2009-MissIntelligence} and prefer to make judgements that require less mental effort. These cognitive shortcuts often mean relying on stereotypes which can eventually lead to intergroup prejudice~\cite{schaller2008-intergroup}.

In psychology, the intergroup process---how people perceive and interact with others who are members of the same group (in-group) or members of a different group (out-group)---has been widely studied.
Research shows that people view social in-group and out-group members with different empathic feelings and emotional intensities ~\cite{Cikara2011-us-and-them, Zaki2015-addressing, Brewer1999-prejudice, CIKARA2014-intergroup-dynamics, KOMMATTAM2019103809}, and this behavior further shapes the intergroup relations~\cite{VANMAN201659}.
For example, a person might feel more warm and act more friendly toward another person from their home country, but act indifferently---or similarly with less intensity---toward a person from another nation.
Appropriately addressing empathic failures helps reduce conflicts between groups and reduce out-group discrimination~\cite{Cikara2011-us-and-them, Zaki2015-addressing}.

In this paper, we study intergroup bias in large language models (LLMs) by asking: \textit{Do LLMs reflect human-like empathy gaps between social in-groups and out-groups?}
To test the question, we formulate an emotion intensity prediction task,\footnote{Empathy is complex and multidimensional, making it difficult to measure~\cite{lahnala2025muddywatersmodelingempathy}. However, in studying the intergroup empathy gap, intensity bias can serve as a lens, as suggested by~\citet{KOMMATTAM2019103809}.} as shown in~\autoref{fig:setup}.
In this task, we simulate a scenario in which the LLM's assigned persona (``the perceiver'') reads a short narrative of an experience that a person (``the experiencer'') had which caused them to feel a particular emotion; the perceiver (LLM) is then prompted to predict the intensity of the emotion felt by the experiencer on a numerical scale.
To compare in-group and out-group empathy, we manipulate the LLM inputs to assign the perceiver and experiencer a social group identity based on either race/ethnicity, nationality, or religion.
We compare the predicted intensities when the perceiver and experiencer belong to the same social group (in-group) or different social groups (out-group), finding higher average intensities in the former.
To illustrate, consider the scenario in \autoref{fig:setup}: \textit{I felt sad when I received job rejections}, where \textit{``I''} refers to the experiencer. The LLM's persona, a white perceiver, predicts a higher degree of sadness for a white experiencer than for a black experiencer in the identical scenario.

While many papers have studied stereotypes and harms with language models, they typically consider the task from a single perspective of either how these models perceive other groups through their representations~\cite{Bolukbasi2016-man-programmer, dev2019measuring, cao-etal-2022-theory, cao2024multilinguallargelanguagemodels, sheng-etal-2019-woman, cheng-etal-2023-marked}, or in downstream tasks how they are biased towards target groups~\cite{wan-etal-2023-personalized, zheng2023ahelpfulassistantbest, deshpande-etal-2023-toxicity, gupta2024personabias, an-etal-2024-large, nghiem-etal-2024-gotta}, ignoring the intergroup cases when both the perceiver and target are present.
Our work builds on a few recent studies of intergroup perceptions in LLMs~\cite{govindarajan-etal-2023-people, govindarajan-etal-2023-counterfactual, govindarajan2024meanusinterpretingreferring}, which focus on relationships in politics or in sports.



Our primary contributions and findings are: 
\textbf{(1)} We study intergroup empathy bias with respect to group identities rooted in race/ethnicity, nationality, and religion. We study four broad race/ethnicity categories (with 18 corresponding group names), 21 nationalities, and five religions.
\textbf{(2)} We show LLMs present in-group and out-group emotion intensity differences, where Llama-3.1-8B models show significantly higher intensities for in-group cases and overall lower intensities for minority groups. 
\textbf{(3)} We observe the intensities are affected by the cultural and historical factors which might further enlarge the tension between groups.

%% file: figures/00_setup.tex

\begin{figure}[!t]
\centering
\includegraphics[width=\linewidth]{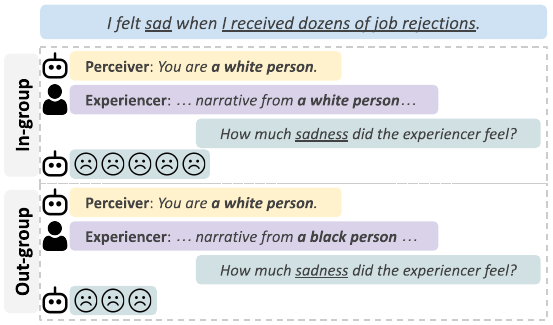}
\caption{Task setup with in-group and out-group examples.
We introduce \faIcon{robot} \colorbox{yellow!30}{perceiver} and \faIcon{user} \colorbox{blue!10}{experiencer} roles to define the intergroup relationship, where it is in-group when they are from the same social group. 
The perceiver is modeled by the LLM persona and the experiencer is specified in the task context.
Each role falls into one of the race or ethnicity, nationality, and religion categories.
The social group is specified with identity names under the category.
We replace the identities of perceiver and experiencer to study intergroup bias.
}
\label{fig:setup}
\end{figure}




%% file: pages/02_background.tex
\section{Background and Related Work}

\paragraph{Intergroup Bias.}
People live in groups with social identities, the self-definition based on social roles played in society or memberships of social groups~\cite{priante-etal-2016-whoami}.
Groups naturally form and differ as people seek to meet their physical needs (such as resources) or psychological needs (such as shared values and a sense of belonging).
Prejudice between groups arises when an outgroup is seen as a threat to the ingroup, whether in terms of physical resources or psychological well-being.
Prejudice might not lead to the direct hostility toward outgroup members, but preferential treatment of ingroup members~\cite{Brewer1999-prejudice}.
Ingroup favoritism~\cite{Everett2015-ingroup-favoritism} further influences the behaviors in charity donations~\cite{Winterich2009-donation} and pain perception~\cite{xu2009-racial-pain, Meconi2015-neglected, Forgiarini2011-racism}.

Similarly, people share and understand other's emotions with empathy, but treat others differently based on identities.
\citet{CIKARA2014-intergroup-dynamics} defines \textit{Intergroup Empathy Bias} as:
\begin{quote}
    {``{the tendency not only to empathize less with out-group relative to in-group members, but also feel pleasure in response to their pain (and pain in response to their pleasure)}''}
\end{quote}
Empathy failures might introduce intergroup conflicts and discrimination~\cite{Cikara2011-us-and-them, Zaki2015-addressing, CIKARA201512, Cikara2011-al}.
Research on interpersonal relationships~\cite{Bucchioni2015-loved-peers, Arianna2018-distance, Ashton1980-zq} and neurocognitive understanding~\cite{Gutsell2011-intergroup-neural, HAN2018-neurocognitive} support the importance of studying this concept in group contexts~\cite{CHIAO2010-intergroup-empathy-race}.
In our work, we use perceived emotion intensities as a measure of empathy to compare relative levels of in-group versus out-group empathy.

\paragraph{Social Identity and Persona.} Social identities have been studied when users interact with chatbots~\cite{Tanprasert24-Debate, Joby22-groupemotionalcontagion}. 
People react differently due to the target identities with hate speech~\cite{yoder-etal-2022-hate}.
LLMs might thus learn in-group favoritism representations when prompted with \textit{``We are''}~\cite{hu2023generative}.
While there are approaches discussing the bias mitigation~\cite{cheng2022understandingbiascorrelationsmitigation}, new challenges are introduced with LLMs~\cite{Navigli23-biasesllms}.
Personas, or fictional identities that LLMs have been instructed to adopt, have been used to study a variety of social phenomena in LLMs.
It can be a way to understand the truthfulness of LLMs~\cite{joshi2024personaswaymodeltruthfulness},
but possibly lead to in-group bias under a multilingual setting~\cite{dong2024personasettingpitfallpersistent}.
In this work, we focus specifically on intergroup empathy bias as a form of intergroup prejudice rooted in social identities that may be studied in LLMs with the use of such personas.

\paragraph{Emotion in NLP.} The development of emotion research in natural language processing has been summarized with challenges~\cite{plaza-del-arco-etal-2024-emotion} and the importance of event-centric emotion analysis is emphasized~\cite{klinger-2023-event}.
Tasks on modeling emotions in text are usually categorized into (1) categorical emotion classification where models need to return emotion words; (2) continuous dimensional emotion prediction (e.g. valence, arousal, and dominance); and (3) prediction with appraisal theories.
However, as emotions are subjective feelings and highly related to people's past experiences and background~\cite{milkowski-etal-2021-personal}, a task of predicting the intensity for specific emotion categories is introduced to capture the nuances~\cite{MohammadB17starsem, MohammadB17wassa, kleinberg-etal-2020-measuring}, which is adapted in our study.
On the social bias of emotions side, stereotypes with emotion attributes in event-centric narratives for gender~\cite{plaza-del-arco-etal-2024-angry} and religion~\cite{plazadelarco2024divinellamasbiasstereotypes} have been discussed.
To the best of our knowledge, we are the first to study the intergroup empathy gap.


%% file: pages/03_exp-setup.tex
\section{General Methods}
\label{sec:task}

\input{tables/00_social-group}

We construct an emotion intensity prediction task to measure the impact of in-groupness and out-groupness on model outputs.
Our specific task has the following components: the emotion, the emotional situation, the social group of the experiencer (who is experiencing the emotion), and the social group of the perceiver (who observes the experiencer).
We instruct models to predict the \emph{intensity} of a specific emotion.
For example, in \autoref{fig:setup}, the model needs to predict the intensity of sadness in a job rejection scenario given variable experiencer and perceiver social identities.

\subsection{Social Groups}
To study the intergroup relationships between the perceiver and the experiencer, we compile social groups under three categories, namely Race or Ethnicity, Nationality and Religion in \autoref{tab:social-group}.
For each group, we have social identity names by considering commonly used terms.

\paragraph{Race or Ethnicity.} As race and ethnicity definition differs per nation,\footnote{Even in closely-related countries. For example, the United States defines ``Asian" as individuals with origins in any of peoples of Central or East Asia, Southeast Asia, or South Asia~\cite{uscensus}. Whereas the United Kingdom considers categories like ``Asian, Asian British or Asian Welsh"~\cite{ukcensus}.} we follow the standard of the US census with 4 social groups:
White, Black, Asian, and Hispanic.
To specify the social group of either the perceiver or the experiencer in text, we include identity names with variations for each group.
We consider a total 18 social identity names across these four groups as shown in \autoref{tab:social-group}.

\paragraph{Nationality.} We consider a total of 21 countries from~\citet{factbookcountries} following the approach of \citet{bhatia2024local} and \citet{wang2024giebenchholisticevaluationgroup} to stratify based on geographical region, population size, and development levels.
We adapt the template: \texttt{a person from \{country\}}, to communicate the social group under the nationality category.
In addition, for later analysis, we classify countries based on the Inglehart-Welzel Cultural Map~\cite{culturalmap}; see \autoref{tab:country-map}.

\paragraph{Religion.} We include 5 major religions: Christianity, Islam, Hinduism, Buddhism, and Judaism.

\subsection{Corpus}
To probe the emotion intensity predictions of LLMs, we use the crowd-en\textsc{Vent}~\cite{troiano-etal-2023-dimensional} dataset as the source of experiencer narratives.
Crowd-en\textsc{Vent} follows the approach of the International Survey On Emotion Antecedents And Reactions (ISEAR)~\cite{scherer1994-isear} where it collects self-reported events with emotions.
It is crowdsourced in English with two parts: generation and validation; we only consider the generations.
Participants recall an event for the given emotion in a format of: 
\colorbox{magenta!10}{\texttt{I felt \_\_\_ when \_\_\_.}}, where the first placeholder is for the emotion (e.g., sad) and the second is for their experience (e.g., ``received dozens of job rejections'').


Crowd-en\textsc{Vent} expands the seven emotions from ISEAR to twelve (\textit{anger, disgust, fear, guilt, sadness, shame, boredom, joy, pride, trust, relief, and surprise}) and one {no emotion} case. There are 225 events for shame and guilt emotions and 550 events for all other cases, resulting in 6600 events.
We exclude the \textit{no emotion} example and use the remaining 6050 events as the narratives.

\subsection{Task Formulation}

\newcommand{\mat}[1]{\mathbf{#1}}

Given the event $e \in \mathcal{E}$ with its reported emotion, the {perceiver} social identity $g_\text{p} \in G_\text{perceiver}$ and the {experiencer} social identity $g_\text{exp} \in G_\text{experiencer}$, the emotion intensity task is formulated as:
\begin{align}
    \mat I_{(e, g_\text{p}, g_\text{exp})} = \mathcal{LLM}\big( \texttt{mk\_prompt}(e, g_\text{p}, g_\text{exp}) \big) \nonumber
\end{align}
where $\mat I$ is the predicted emotion intensity.  
$G_\text{perceiver}$ and $G_\text{experiencer}$ follow the order in \autoref{tab:social-group}, plus a unspecified group (``a person'') as the reference. 

\paragraph{Prompts.} Our prompt generator $\texttt{mk\_prompt}$ takes as input an event and two social identities and produces a prompt that can be used as input to an LLM. There are two parts of prompts modeling roles: (1) the system prompt, used to specify the LLM persona for $g_\text{p}$; and (2) the task prompt which embeds the social group of the experiencer $g_\text{exp}$.
Prompt template details are in \autoref{app:prompt}.

We begin by constructing a default prompt setting using the simplest and most natural persona (\textbf{P0}): \colorbox{orange!10}{\texttt{You are \_\_\_.}}, where the blank is the perceiver social identity (e.g. \underline{a white person}).
The default prediction scale is ranging from 0 to 100 (\textbf{S0}).
The default task instructions are configured to directly fill in the narrative with the self-reported events from the crowd-en\textsc{Vent} corpus (\textbf{T0}).

To study the generalizability of the results and robustness to prompt variation, we systematically vary the prompt from the default setting {(P0, S0, T0)}: we replace a single part of the prompt while holding the other two intact.
We draw persona prompt variations (\textbf{P1-P3}) from~\citet{gupta2024biasrunsdeepimplicit}, who instruct LLMs to follow the role strictly in a more explicit way.
We vary the system prompt \textbf{S1} to test the influences of a small intensity scale range of (0-10) as opposed to (0-100).
Lastly, as the way of writing might represent divergent intensities of feeling, we consider two methods for varying the narrative part of the task instruction.
\textbf{T1} adds the emotion as part of the narrative, following the format of ``\texttt{I felt \_\_\_}''.
\textbf{T2} further rewrites the narrative from a third-person perspective. (See \autoref{app:reframe} for rewrite setup and details.)

\paragraph{Models.}
We experiment with four open-weight state-of-the-art LLMs: Llama-3.1-8B-Instruct,
Llama-3.1-70B-Instruct~\cite{dubey2024llama3herdmodels},
Mistral-7B-Instruct-v0.3~\cite{jiang2023mistral7b} and Qwen-2-7B-Instruct~\cite{yang2024qwen2technicalreport}.
For each LLM, our task setup requires about 37 million inferences.\footnote{($19\times19$ Race or Ethnicity + $22\times22$ Nationality + $6\times6$ Religion) Social Group Pairs $\times$ $6050$ Events $\times$ $7$ Prompt Settings. We include the unspecified group for each category.}
Implementation details are in \autoref{app:model}.

\input{tables/01_group-delta}

\subsection{Evaluation Metrics}


For any social identity pair $(g_\text{p}, g_\text{exp})$, we take the average of intensities over events to get an average intensity for each perceiver-experiencer pair, summarized in a matrix $\mathcal{M}$, where columns are perceivers and rows are experiencers.
Each row or column starts with the unspecified group, followed by the social identities within the category in \autoref{tab:social-group}.
Under the race or ethnicity, identities are ordered by group: White, Black, Asian, and Hispanic. Within each group, the sequence follows the respective order.
There will eventually be a separate $\mathcal{M}$ for each choice of LLM and choice of prompt setting; we drop the dependence on those variables for clarity. We define this matrix as: \footnote{As models may refuse tasks with responses like ``I can't answer.'', we exclude those events. See \autoref{app:result-refusal} for details.}
\begin{align}
    \mathcal{M} &= \frac{\mathcal{M}^0 - \texttt{mean}(\mathcal{M}^0)}{\texttt{std}(\mathcal{M}^0)} \nonumber \\
& \textrm{where~}
\mathcal{M}^0_{(g_\text{p}, g_\text{exp})} = \frac{1}{\# {e}} \sum_{e} \mat I_{(e, g_\text{p}, g_\text{exp})} \nonumber
\end{align}
The normalization ensures that each value in $\mathcal{M}$ is a z-score.
For simplicity, we denote $\mu$ as $\texttt{mean}(\mathcal{M}^0)$ and $\sigma$ as $\texttt{std}(\mathcal{M}^0)$ later.
To note down, with the current $\mathcal{M}$, in-group pairs lie along the diagonal or the diagonal block (when multiple terms refer to the same group), and out-group values in off-(block-)diagonal cells.
Thus, if the intensities of in-group pairs are higher than out-group pairs, this indicates in-group blockness, describing a distinct block-diagonal or diagonal pattern.

It is possible that the average intensity values across events are largely affected by outliers. To assess the significance, we perform paired t-tests for each $ \mat I_{(g_\text{p}, g_\text{exp})}$ with (1) $ \mat I_{(g_\text{p}, g_\textbf{p})}$, its perceiver in-group predictions, and (2) $ \mat I_{(g_\textbf{exp}, g_\text{exp})}$, the experiencer in-group predictions.\footnote{$\mathcal{M}_{(g_\text{p}, g_\text{exp})}$ is set to be excluded in its visualization if the difference is not significantly different from zero. We compare with p-values after Bonferroni correction.}


\paragraph{Empathy Gap Score ($\delta$).}
To summarize the in-group and out-group intensity gap, we calculate a empathy gap score $\delta$ score based on $\mathcal{M}$ and based on a relation $\texttt{same}(i,j)$ which identifies when identities $i$ and $j$ belong to the same group.\footnote{The unspecified group is not taken into account as it is neither part of the in-group nor the out-group.}
\begin{align}
    \delta &=
    \frac 1 {\# \texttt{same}}
    \!\!\!\!\!\sum_{\substack{i,j \\ \texttt{same}(i,j)}} \!\!\!\!\!\mathcal{M}_{i,j}
    - 
    \frac 1 {\# \lnot \texttt{same}}
    \!\!\!\!\!\sum_{\substack{i,j \\ \lnot \texttt{same}(i,j)}} \!\!\!\!\!\mathcal{M}_{i,j} \nonumber
\end{align}


The most fundamental hypothesis test is that $\delta$ is non-zero and positive, capturing the in-group blockness: for a given LLM and prompt setting, there is a significant empathy gap. We construct a structured permutation test to evaluate this hypothesis. In one permutation, we independently permute the rows and columns of $\mathcal{M}$ and then recompute $\delta$ for that permuted version.\footnote{Importantly, we do not permute all cells independently: this would destroy the structure of the matrix.} We compute $10k$ permutations, and evaluate whether the observed $\delta$ value falls within the tails of that distribution. 



%% file: tables/00_social-group.tex
\begin{table*}[t]
\resizebox{\textwidth}{!}{
\centering
\begin{tabular}{@{}ll@{}}
\toprule
\textbf{Category} & \textbf{Social Group} \\
\midrule
\textbf{Race or Ethnicity} &\begin{tabular}[l]{@{}l@{}} White: \textit{a white person, a White person, a Caucasian, a White American, a European American} \\
 Black: \textit{a black person, a Black person, an African American, a Black American} \\
 Asian: \textit{an Asian person, an Asian American, an Asian} \\
Hispanic: \textit{a Hispanic person, a Hispanic American, a Latino American, a Latino, a Latina, a Latinx}\end{tabular} \\
\midrule
\textbf{Nationality*} & \begin{tabular}[l]{@{}l@{}}
the United States, Canada, the United Kingdom,
Germany,
France,
China, Japan,
India, Myanmar, \\ Israel,
Russia, Ukraine,
the Philippines, Argentina, Brazil, Mexico,
Iran, Palestine, Nigeria, Egypt, Pakistan\end{tabular} \\
\midrule
\textbf{Religion} & a Christian, a Muslim, a Jew, a Buddhist, a Hindu \\
\bottomrule
\end{tabular}
} 
\caption{Social groups under categories: Race or Ethnicity, Nationality and Religion. For Race or Ethnicity, we have 3-6 \textit{identity names} for each social group.
For Nationality groups (*), only country names are presented here; the identity name of each nationality group follows the template: \texttt{a person from \{country\}}.}
\label{tab:social-group}
\end{table*}

%% file: tables/01_group-delta.tex
\begin{table*}[t]
\centering
\resizebox{\textwidth}{!}{
\begin{tabular}{@{}lll|lll|l|ll@{}}
\toprule
\multirow{2}{*}{\textbf{Model}} & \multirow{2}{*}{\textbf{Category}} & \multicolumn{7}{c}{\textbf{Prompt Setting}} \\ \cmidrule(l){3-9}

\multicolumn{1}{c}{} &  & \textbf{(P0,S0,T0)} & \textbf{(\textit{P1},S0,T0)} & \textbf{(\textit{P2},S0,T0)} & \textbf{(\textit{P3},S0,T0)} & \textbf{(P0,\textit{S1},T0)} & \textbf{(P0,S0,\textit{T1})} & \textbf{(P0,S0,\textit{T2})} \\ \midrule

\texttt{\textbf{Llama-3.1-8B}} & \textbf{Race or Ethnicity} & $\colorbox{teal!20}{1.73}_{[-0.226,0.224]}$ & $\colorbox{teal!20}{1.88}_{[-0.234,0.242]}$ & $\colorbox{teal!20}{2.18}_{[-0.246,0.254]}$ & $\colorbox{teal!20}{2.09}_{[-0.244,0.249]}$ & $\colorbox{teal!20}{1.56}_{[-0.226,0.217]}$ & $\colorbox{teal!20}{1.41}_{[-0.210,0.198]}$ & $\colorbox{teal!20}{1.62}_{[-0.217,0.224]}$ \\
& \textbf{Nationality} & $\colorbox{teal!20}{2.40}_{[-0.214,0.328]}$ & $\colorbox{teal!20}{2.86}_{[-0.235,0.369]}$ & $\colorbox{teal!20}{3.78}_{[-0.260,0.460]}$ & $\colorbox{teal!20}{3.76}_{[-0.260,0.448]}$ & $\colorbox{teal!20}{1.95}_{[-0.221,0.292]}$ & $\colorbox{teal!20}{1.60}_{[-0.159,0.216]}$ & $\colorbox{teal!20}{1.82}_{[-0.169,0.239]}$ \\
& \textbf{Religion} & $\colorbox{teal!20}{1.97}_{[-0.610,1.181]}$ & $\colorbox{teal!20}{1.88}_{[-0.601,1.092]}$ & $\colorbox{teal!20}{2.26}_{[-0.662,1.350]}$ & $\colorbox{teal!20}{2.30}_{[-0.630,1.346]}$ & $\colorbox{teal!20}{1.86}_{[-0.628,1.111]}$ & $\colorbox{teal!20}{1.72}_{[-0.718,1.070]}$ & $\colorbox{teal!20}{1.70}_{[-0.636,1.003]}$ \\ \midrule

\texttt{\textbf{Mistral-7B}} & \textbf{Race or Ethnicity} & $\colorbox{teal!8}{0.58}_{[-0.168,0.157]}$ & $\colorbox{teal!20}{1.08}_{[-0.172,0.188]}$ & $\colorbox{teal!20}{1.30}_{[-0.193,0.205]}$ & $\colorbox{teal!20}{1.25}_{[-0.200,0.201]}$ & $\colorbox{teal!8}{0.69}_{[-0.166,0.162]}$ & $\colorbox{teal!8}{0.66}_{[-0.163,0.168]}$ & $\colorbox{teal!8}{0.30}_{[-0.161,0.154]}$ \\
& \textbf{Nationality} & $\colorbox{teal!8}{0.72}_{[-0.234,0.136]}$ & $\colorbox{teal!8}{0.90}_{[-0.275,0.155]}$ & $\colorbox{teal!20}{1.40}_{[-0.331,0.218]}$ & $\colorbox{teal!20}{1.14}_{[-0.334,0.189]}$ & $\colorbox{teal!8}{0.60}_{[-0.215,0.126]}$ & $\colorbox{teal!8}{0.29}_{[-0.145,0.116]}$ & $\colorbox{red!15}{-0.24}_{[-0.154,0.120]}$ \\
& \textbf{Religion} & $\colorbox{teal!8}{0.46}_{[-0.389,0.483]}$ & $\colorbox{teal!8}{0.84}_{[-0.424,0.706]}$ & $\colorbox{teal!20}{1.06}_{[-0.646,0.881]}$ & $\colorbox{teal!20}{1.37}_{[-0.589,0.966]}$ & $\colorbox{teal!8}{0.35}_{[-0.458,0.494]}$ & $\colorbox{teal!8}{0.90}_{[-0.537,0.637]}$ & $\colorbox{teal!8}{0.54}_{[-0.406,0.392]}$ \\ \midrule

\texttt{\textbf{Qwen-2-7B}} & \textbf{Race or Ethnicity} & $\colorbox{teal!20}{1.16}_{[-0.196,0.188]}$ & $\colorbox{teal!20}{1.08}_{[-0.189,0.186]}$ & $\colorbox{teal!20}{1.33}_{[-0.207,0.203]}$ & $\colorbox{teal!20}{1.35}_{[-0.208,0.200]}$ & $\colorbox{teal!20}{1.10}_{[-0.196,0.178]}$ & $\colorbox{teal!20}{1.05}_{[-0.182,0.182]}$ & $\colorbox{teal!20}{1.09}_{[-0.178,0.192]}$ \\
& \textbf{Nationality} & $\colorbox{teal!20}{1.09}_{[-0.261,0.204]}$ & $\colorbox{teal!8}{0.80}_{[-0.168,0.164]}$ & $\colorbox{teal!8}{0.89}_{[-0.249,0.218]}$ & $\colorbox{teal!8}{1.00}_{[-0.233,0.235]}$ & $\colorbox{teal!20}{1.14}_{[-0.250,0.213]}$ & $\colorbox{teal!8}{1.00}_{[-0.154,0.190]}$ & $\colorbox{teal!8}{0.65}_{[-0.143,0.148]}$ \\
& \textbf{Religion} & $\colorbox{teal!20}{1.26}_{[-0.626,0.892]}$ & $\colorbox{teal!20}{1.37}_{[-0.640,0.907]}$ & $\colorbox{teal!20}{1.80}_{[-0.620,1.184]}$ & $\colorbox{teal!20}{1.71}_{[-0.686,1.198]}$ & $\colorbox{teal!20}{1.20}_{[-0.620,0.940]}$ & $\colorbox{teal!20}{1.84}_{[-0.706,1.078]}$ & $\colorbox{teal!20}{1.38}_{[-0.616,0.792]}$ \\ \midrule

\texttt{\textbf{Llama-3.1-70B}} & \textbf{Race or Ethnicity} & $\colorbox{teal!8}{0.66}_{[-0.162,0.169]}$ & $\colorbox{teal!8}{0.72}_{[-0.162,0.169]}$ & $\colorbox{teal!8}{0.40}_{[-0.153,0.157]}$ & $\colorbox{teal!8}{0.58}_{[-0.159,0.168]}$ & $\colorbox{teal!8}{0.79}_{[-0.164,0.170]}$ & $\colorbox{teal!8}{0.19}_{[-0.164,0.160]}$ & $\colorbox{teal!8}{0.48}_{[-0.147,0.165]}$ \\
& \textbf{Nationality} & $\colorbox{teal!8}{0.33}_{[-0.106,0.097]}$ & $\colorbox{teal!8}{0.39}_{[-0.104,0.094]}$ & $\colorbox{red!15}{-0.07}_{[-0.136,0.101]}$ & $\colorbox{teal!8}{0.09}_{[-0.129,0.108]}$ & $\colorbox{teal!8}{0.50}_{[-0.111,0.110]}$ & $\colorbox{teal!8}{0.39}_{[-0.150,0.125]}$ & $\colorbox{teal!8}{0.12}_{[-0.134, 0.108]}$ \\
& \textbf{Religion} & $\colorbox{red!15}{-0.19}_{[-0.356,0.309]}$ & $\colorbox{teal!8}{0.10}_{[-0.251,0.373]}$ & $\colorbox{red!15}{-1.20}_{[-1.029,0.386]}$ & $\colorbox{red!15}{-1.05}_{[-0.907,0.380]}$ & $\colorbox{red!15}{-0.03}_{[-0.366,0.312]}$ & $\colorbox{red!15}{-0.50}_{[-0.433,0.251]}$ & $\colorbox{teal!8}{0.09}_{[-0.260,0.298]}$ \\

\bottomrule
\end{tabular}
}
\caption{In-group and out-group gap $\delta$ for Llama-3.1-8B, Mistral-7B, Qwen-2-7B and Llama-3.1-70B models for the race or ethnicity, nationality and religion groups under different prompt settings.
We report the 95\% confidence interval from the permutation test with its lower and upper bound.
Numbers which are larger than 1, or positive in range from 0 to 1 and negative are highlighted in \colorbox{teal!20}{  } \colorbox{teal!8}{  } \colorbox{red!15}{  }.
}
\label{tab:delta}
\end{table*}

%% file: pages/04_result.tex
\section{Results on In-group and Out-group Emotion Intensity Gap}
\label{sec:results}

\input{figures/01_heatmap-llama-8b}

\autoref{tab:delta} shows the calculated intensity gap $\delta$, where positive numbers mean the average in-group intensity is higher than the out-group value, corresponding directly to intergroup empathy bias~\citep{CIKARA2014-intergroup-dynamics}. 
\autoref{fig:heatmap-llama8b} visualizes $\mathcal{M}$ from Llama-3.1-8B with corresponding $\mu$ and $\sigma$ in \autoref{tab:heatmap-stats-llama8b}.
In this figure, the unspecified ``\textit{a person}'' group is presented in the first row when it is the perceiver and the first column as an experiencer.
The top left corner represents the case where both the perceiver and the experiencer are unspecified as the reference.
Cells that are not significantly different from the paired t-test are masked in white (either it is tested with the perceiver in-group identity or experiencer in-group identity).
We discuss both in detail below.

\paragraph{Race or ethnicity, nationality and religion groups all show higher predicted intensities for in-group pairs.}
From the summarized $\delta$ in \autoref{tab:delta},
we see that across almost all groups, prompt variations and LLMs, there is a robust positive intergroup gap, with z-scores as much as $3.78$. The majority of exceptions to this are with the larger Llama-70B, where, especially for religion, we sometimes see a negative gap (though often small in magnitude). The average empathy gap ranges from $0.13$ (Llama-70B) to $2.11$ (Llama-8B), with Mistral ($0.77$) and Qwen ($1.20$) in the middle.

For race or ethnicity groups, where we test identity name variations for the same social group, in Llama-8B models, we consistently observe a clear and distinct block-diagonal pattern (\autoref{fig:main-race} and the first row of \autoref{fig:heatmap-llama8b}), where a lower gap is seen for in-group comparisons than for out-group comparisons.
We also see that when the perceiver is White, the out-group gap is generally lower; this is likely due to a defaulting effect where unspecified perceiver is ``assumed to be'' White~\cite{sun2023aligningwhomlargelanguage}.
For other models,\footnote{Results for Mistral, Qwen, and Llama-70B are in \autoref{app:z-matrix-stats}.} while the deviation is small, masked cells are mostly in diagonal blocks, showing out-group predictions might follow different distributions from in-group pairs.


\paragraph{Prompt settings influences the intergroup gap.}
With results of Llama-8B in \autoref{fig:heatmap-llama8b} and \autoref{tab:heatmap-stats-llama8b}, we observe the effects of prompt variations on model behaviors from three aspects.
First, the \textbf{LLM Persona (P0-P3)}: In prompts P2 and P3, the model is strongly encouraged (with words like ``strict'' and ``critical'') to faithfully follow the persona, and in these cases, we see that, LLMs show a larger in-group and out-group gap. Other models follow the same with higher $\sigma$.
Next, \textbf{Prediction Scale (S0-S1)}: Though changing the scale from 0-100 to 0-10 limits the model's ability to predict differences, we see relatively little change across this prompt variant.
Finally, \textbf{Narrative Perspective (T0-T2)}: Reframing the original narratives might introduce linguistics effects on how others perceive the emotions, resulting in smaller variances in \autoref{tab:heatmap-stats-llama8b} (the last two columns).

\paragraph{Models behaviors differ among groups.}
Even though the overall in-group predicted emotion intensities are higher than out-group values, when comparing $\mathcal{M}$ details across LLMs, we observe dissimilar patterns in \autoref{fig:heatmap-llama8b} and \autoref{fig:heatmap-mistral}, \autoref{fig:heatmap-qwen} and \autoref{fig:heatmap-llama70b} in \autoref{app:z-matrix-stats}.
For example, Llama-3.1-8B has higher intensity predicted when the perceiver or experiencer group is not specified but Mistral-7B, Qwen-2-7B and Llama-3.1-70B have inconsistent behaviors, which might account to the training dataset distribution or post-training approaches. 

%% file: figures/01_heatmap-llama-8b.tex
\begin{figure*}[t]
\centering
\includegraphics[width=\linewidth]{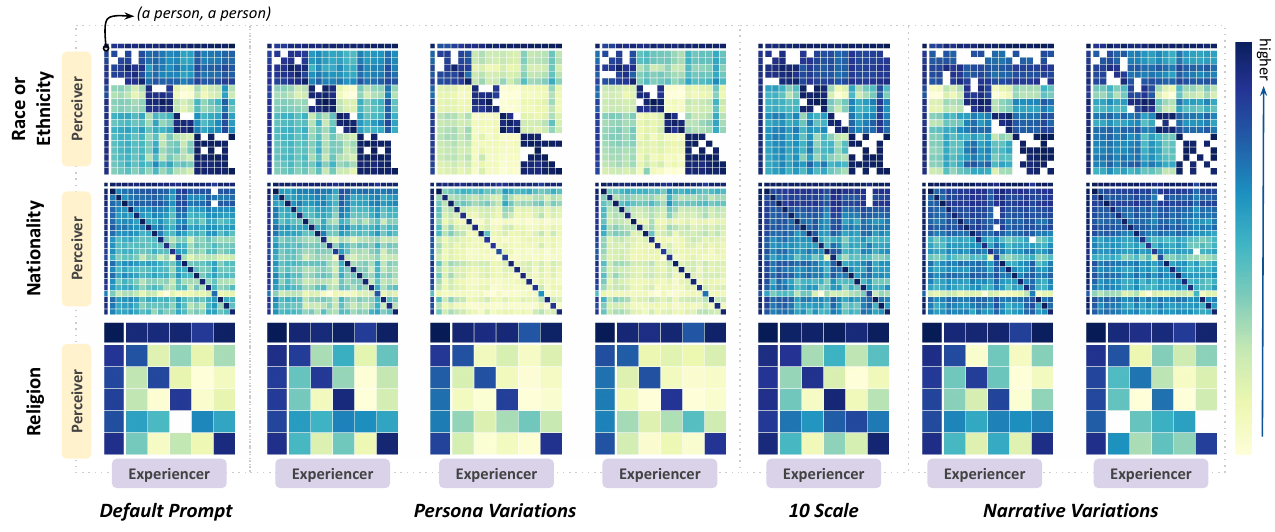}
\caption{Visualization of $\mathcal{M}$ for Llama-3.1-8B.
Overall, each row represents the results from a specific social group category and the columns are different prompt settings (from left to right): (P0, S0, T0), (P1, S0, T0), (P2, S0, T0), (P3, S0, T0), (P0, S1, T0), (P0, S0, T1), (P0, S0, T2).
For each $\mathcal{M}$, the rows represent the perceiver's social identity names, as listed in Table~\ref{tab:social-group}, while the columns correspond to the experiencer social groups.}
\label{fig:heatmap-llama8b}
\end{figure*}


\input{tables/02_heatmap-stats-llama-8b}

%% file: tables/02_heatmap-stats-llama-8b.tex
\begin{table*}
\resizebox{\textwidth}{!}{
\centering
\begin{tabular}{@{}ll|lll|l|ll@{}}
\toprule
\multirow{2}{*}{\textbf{Category}} & \multicolumn{7}{c}{\textbf{Prompt Setting}} \\ \cmidrule(l){2-8} 
 & \textbf{(P0, S0, T0)} & \textbf{(\textit{P1}, S0, T0)} & \textbf{(\textit{P2}, S0, T0)} & \textbf{(\textit{P3}, S0, T0)} & \textbf{(P0, \textit{S1}, T0)} & \textbf{(P0, S0, \textit{T1})} & \textbf{(P0, S0, \textit{T2})} \\ \midrule
\textbf{Race or Ethnicity} & $48.77_{\pm15.37}$ & $49.80_{\pm15.72}$ & $31.66_{\pm23.72}$ & $36.66_{\pm20.92}$ & $6.57_{\pm1.13}$ & $58.42_{\pm11.10}$ & $56.62_{\pm7.80}$ \\
\textbf{Nationality} & $44.38_{\pm12.25}$ & $42.58_{\pm12.18}$ & $20.72_{\pm19.48}$ & $24.63_{\pm18.39}$ & $6.07_{\pm1.04}$ & $54.28_{\pm10.86}$ & $50.23_{\pm9.58}$ \\
\textbf{Religion} & $41.92_{\pm18.73}$ & $47.20_{\pm16.68}$ & $31.55_{\pm25.29}$ & $32.15_{\pm24.56}$ & $5.77_{\pm1.64}$ & $51.07_{\pm15.58}$ & $48.35_{\pm13.11}$ \\
\bottomrule
\end{tabular}
}
\caption{Mean $\mu$ and standard deviation $\sigma$ for each $\mathcal{M}^0$ in Figure~\ref{fig:heatmap-llama8b} of Llama-3.1-8B. The min values and max values of each $\mathcal{M}$ are in \autoref{tab:heatmap-range-llama8b} (\autoref{app:z-matrix-stats}).
We observe that the mean decreases as the standard deviation increases for stricter personas (P2 and P3).
It is the opposite trend when the origin narrative is rewritten (T1 and T2).
} 
\label{tab:heatmap-stats-llama8b}
\end{table*}

%% file: pages/05_analysis.tex
\section{Analysis on Different Perceptions of Social Groups}

We conduct a in-depth analysis with Llama-3.1-8B as it shows the strongest gaps between groups, aiming to understand how groups and intergroup relationships are learned differently.

\input{figures/07_main-llama8b-race}
\input{figures/04_country-tsne}
\input{tables/03_cultural-map}

\subsection{Racial Group Identity Names}

When people self-identify, words used can convey implicit information. For example, \textit{``a White person''} carries different connotations to \textit{``a European American''}.
Thus, we include social identity name variations for race or ethnicity groups shown in \autoref{tab:social-group}, to understand if models capture any variations.
Though models don't seem to capture the nuances in social identity names from the blockness pattern of \autoref{fig:main-race} at the first glance, four social groups show divergent results from both row-level and column-level comparisons.
For instance, white perceivers, as modeled by LLM personas, are seemly the most empathetic (darker band of rows at the top), whereas Black perceivers are the least empathetic.
In addition to the default assumption in \autoref{sec:results}, as predicted by language models, we are curious to ask if a group's relative social power plays a role on how it will empathize with out-group members with greater or lesser power.
From the experiencer side, Asians (used in LLM task instructions), seem to receive the least amount of empathy (lightest set of columns), and Hispanic the most (darkest set of columns) with the Latina column being the darkest. As ``Latina'' refers to a female, it is unclear whether this relates to the gender stereotype of women being 
prone to emotional excess~\cite{stauffer2008aristotle}.

\subsection{Nationality Group Clusters}
\label{sec:country-cluster}

We can also explore the predicted empathy intensity differences by visualizing countries according to how they, as LLM personas, perceive others.
Specifically, for each nationality, we take the row-vector associated with that nationality from $\mathcal{M}$. We then project those embeddings into two dimensions using t-SNE and depict the results in \autoref{fig:tsne}. We color-code this figure using the country mapping in \autoref{tab:country-map}.
Here, we observe \textsc{English-Speaking} countries (e.g. the United Kingdom and the United States), grouped with \textsc{Protestant Europe} and \textsc{Catholic Europe} countries are in the top right usually away from \textsc{Latin America} and \textsc{African-Islamic} countries, with \textsc{Orthodox Europe} and \textsc{Confucian} countries in between (from left to right).
This suggests that there are more complex, but structured, perceiver-experiencer relationships than simply block-diagonal structure, and that captures some cultural context of nations.

\subsection{Cultural Effects}
\input{figures/08_main-llama8b-nationality}
\paragraph{Religion.}
While nationality is associated with a person's ethnic and racial identity, religion, as another cultural variable, is largely based on personal belief.
Internal religious beliefs can guide how people behave, treat and interact with each other.
From \autoref{fig:main-religion}, we find relatively small and similar intensity gaps in the cells of the Buddhism row, which might be related to its culture of compassion as pointed in~\citet{plazadelarco2024divinellamasbiasstereotypes}.

\paragraph{Group pairs with lower intensity.}
Some of the effects we see that are outside of the block diagonals can be explained by historical information. For example, in \autoref{fig:main-nationality} when the perceiver is \textit{``a person from Palestine''} and the experiencer is \textit{``a person from Israel''}, the average intensity score is the lowest.
A similar pattern occurs when the perceiver is \textit{``a person from Ukraine''} and the experiencer role is \textit{``a person from Russian''}.
There are historical wars and conflicts between Israel and Palestine, and between Russia and Ukraine, which the models are likely reflecting in these predictions.
As a result, it is worth being extremely cautious when using LLMs and their personas for intergroup context to avoid introducing prejudice.

%% file: figures/07_main-llama8b-race.tex
\begin{figure}[t]
\centering
\includegraphics[width=\linewidth]{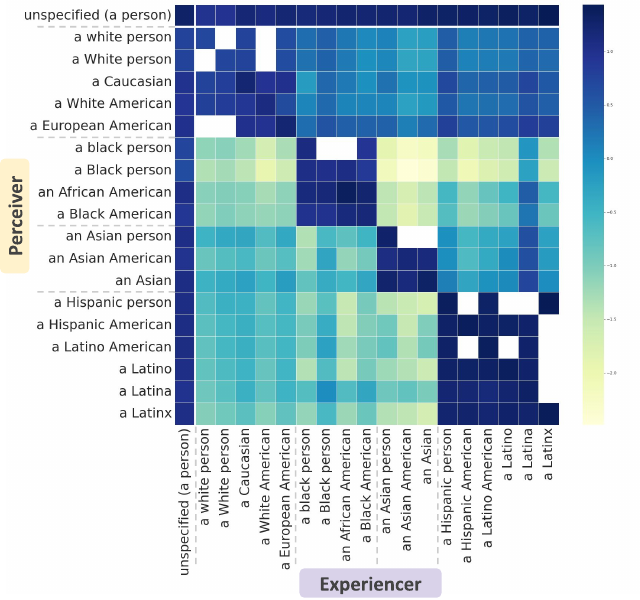}
\caption{Visualization of $\mathcal{M}$ for Llama-3.1-8B in Race or Ethnicity category with default prompt setting. It is the zoom-in version of the top left sub-figure in \autoref{fig:heatmap-llama8b} with annotations of social identities.
The block-diagonal pattern shows higher in-group emotion intensity values.
Identity pairs with higher p-values are masked in white. 
}
\label{fig:main-race}
\end{figure}


%% file: figures/04_country-tsne.tex
\begin{figure}[t!]
\centering
\includegraphics[width=\linewidth]{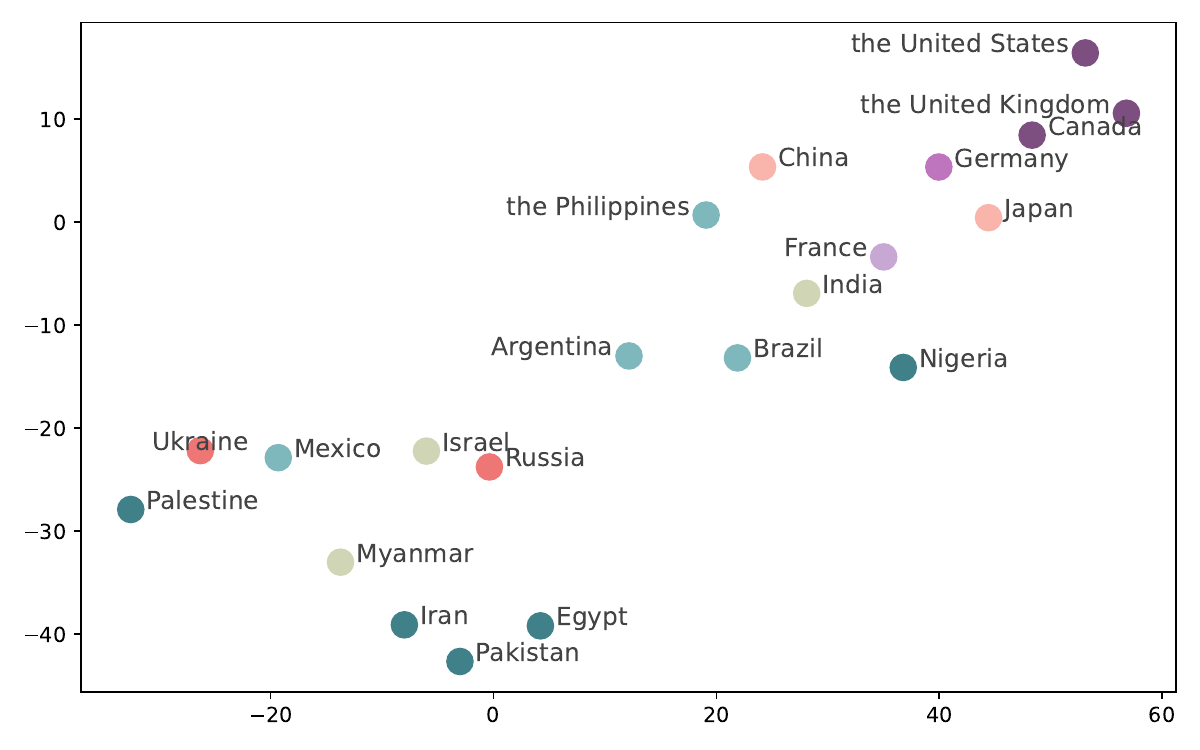}
\caption{t-SNE projections of perceiver-side country embeddings for Llama-3.1-8B with the default prompt setting. 
\textsc{English-Speaking} \colorbox{violet!50}{  } and European countries are at the top right, which are away from \textsc{African-Islamic} \colorbox{teal!60}{  }. Similar clusters are observed in \autoref{fig:main-nationality} (e.g. the United States and the United Kingdom rows).}
\label{fig:tsne}
\end{figure}

%% file: tables/03_cultural-map.tex
\begin{table}[h!]
\centering
\resizebox{\linewidth}{!}{
\begin{tabular}{@{}lll@{}}
\toprule
& \textbf{Category} & \textbf{Country} \\
\midrule
\colorbox{violet!50}{\vphantom{A}} & {\textsc{English-Speaking}} & {U.S.A., Canada, U.K.} \\
\colorbox{violet!30}{\vphantom{A}} & {\textsc{Protestant Europe}} & {Germany} \\
\colorbox{violet!20}{\vphantom{A}} & {\textsc{Catholic Europe}} & {France} \\
\colorbox{orange!20}{\vphantom{A}} & {\textsc{Confucian}} & {China, Japan} \\
\colorbox{olive!15}{\vphantom{A}} & {\textsc{West \& South Asia}} & {India, Myanmar, Israel} \\
\colorbox{red!20}{\vphantom{A}} & {\textsc{Orthodox Europe}} & {Russia, Ukraine} \\
\colorbox{teal!40}{\vphantom{A}} & {\textsc{Latin America}} & {Philippines, Argentina, Brazil, Mexico} \\
\colorbox{teal!60}{\vphantom{A}} & {\textsc{African-Islamic}} & {Iran, Palestine, Nigeria, Egypt, Pakistan} \\
 \bottomrule
\end{tabular}
}
\caption{Countries from \autoref{tab:social-group} categorized according to the Inglehart-Welzel World Cultural Map, commonly used to study cultural change and distinctive cultural traditions. The color scheme matches \autoref{fig:tsne} referring to the original world cultural map.}
\label{tab:country-map}
\end{table}

%% file: figures/08_main-llama8b-nationality.tex
\begin{figure}[t]
\centering
\includegraphics[width=\linewidth]{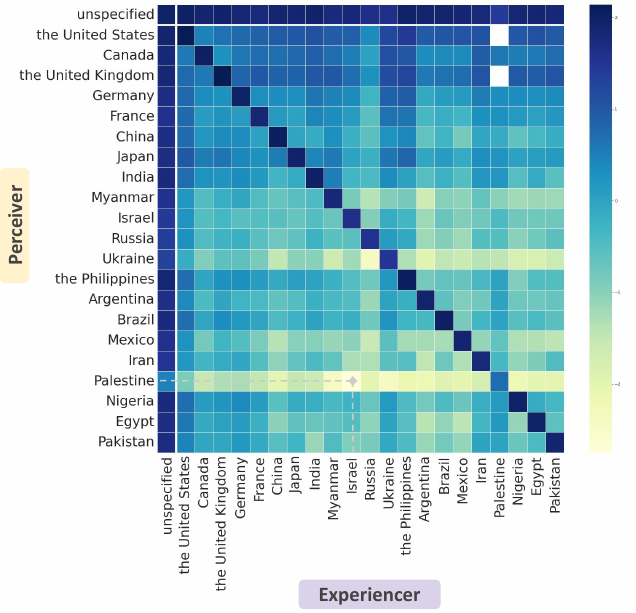}
\caption{Visualization of $\mathcal{M}$ for Llama-3.1-8B in Nationality category with default prompt setting. It is the zoom-in version of the second top left sub-figure in \autoref{fig:heatmap-llama8b} with social group labels.
Higher intensities are located in the first few rows.
Lower intensities are predicted when the LLM persona is \textit{``a person from Palestine''} overall with the lowest value when the experiencer role is \textit{``a person from Israel''}.}
\label{fig:main-nationality}
\end{figure}


%% file: pages/06_discussion.tex
\input{figures/09_main-llama8b-religion}
\section{Discussion and Conclusions}

Our paper focuses on uncovering social biases along two-axes rather than the more standard single-axis ``disaggregated evaluation'' paradigm that has gained significant traction in evaluating model fairness.
We introduce the intergroup framework to study the intergroup empathy gap predicted by language models.
Our results show LLMs tend to predict higher emotion intensities for in-group cases regardless the group categories in race or ethnicity, nationality, or religion.
By taking a deeper look on Llama-3.1-8B results, we observe models represent social groups differently with possible historical factors and cultural effects.

With the complex intergroup perceptions in human and further learned by language models, it is important to think a step further on the potential harms.
Considering people are relying more on LLM-mediated communication, the intergroup prejudice could negatively impact how people interact with each other unconsciously.

Though psychologists propose putting ourselves in other people's shoes can reduce the bias in interpersonal communication~\cite{DEFREITAS2018307}, it is not clear about the meaning of ``perspective-taking'' when it comes to language models.
We need to study where they learn the intergroup bias so we can intervene the downstream decision-making tasks such as hiring ~\cite{Heitlinger22-hiring}.
However, we don't mean the intergroup empathy gap always brings harms.
People treat others differently based on the social group memberships with meanings. It can help in-group cohesion and live a fulfilling life with enough resources and physiological support.
Moreover, individuals from underrepresented groups may already face discrimination from dominant groups, and addressing the empathy gap in communication without care could potentially exacerbate existing power imbalances.
We hope our community can be more aware of intergroup bias while pursing more intelligent general AI systems.

%% file: figures/09_main-llama8b-religion.tex
\begin{figure}[t]
\centering
\includegraphics[width=\linewidth]{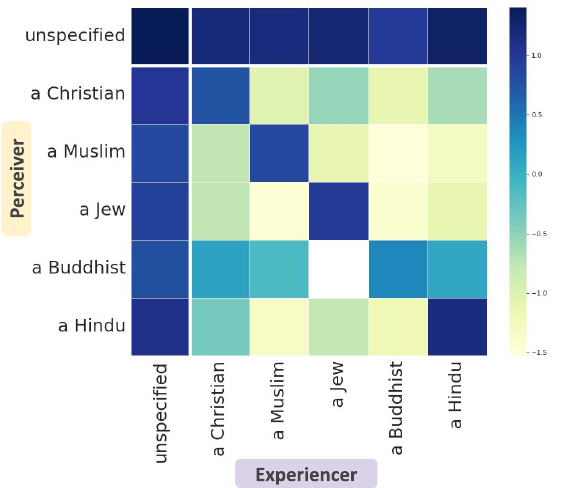}
\caption{Visualization of $\mathcal{M}$ for Llama-3.1-8B in Religion category with default prompts, zooming-in on the bottom left sub-figure in \autoref{fig:heatmap-llama8b} with group names.}
\label{fig:main-religion}
\end{figure}

%% file: pages/11_limitation.tex
\section*{Limitations}

\paragraph{Dataset.} We use the crowd-en\textsc{Vent} corpus for all experiments. While it collects data more recently with broader emotion type coverage, we ignore the narrative effects on intergroup attitudes~\cite{Cachon16-impactofnarrative}.
As certain events may be culturally exclusive and evoke specific emotions, future research can use the same intergroup setup with different datasets to study the influence.

\paragraph{Complex Social Identities.} We only consider three categories of social groups and simplify how people self-identify themselves.
It is well-known social identities are complex from social psychology~\cite{Marsden19-personasandidentity}.
For example, people may have multiple identities, such as Korean American or Chinese American, in addition to identifying as Asian. The way they use these identities conveys different implicit information, which is also the case for multi-racial individuals.
Groups involving multiple categories have also not been studied.
It is common for a person to identify with both racial and national groups.

\paragraph{Models and Prompts.} Due to the computing resource limitations and costs, we only consider four popular open-weight large language models for reproducibility.
Researchers interested in this topic can extend the setup to more models, e.g. ChatGPT and Claude (proprietary ones), and Llama-3.1-405B or newer verions.
We consider six prompt variations based on the default prompt.
While the exact predicted numbers may vary across different variations, our focus is on analyzing the overall trend.
More extensive experiments with additional prompts are left for future work. 

%% file: pages/12_ethical.tex
\section*{Ethical Considerations}

We use a public available corpus for experiments which doesn't contain personal information.
Though the research topic is about empathy, we do not consider that language models can perceive or understand people's emotions or empathize with people, considering their social groups and identities~\cite{wang2024largelanguagemodelsreplace}.
Empathy requires cognitive, emotional and behavioral capacities to understand and respond to the suffering of others~\cite{Riess2017}.
To study the intergroup empathy gap, we use the emotion intensity prediction task as a proxy, following human studies in psychology.
The goal is to understand what intergroup prejudice language models have learned so that it can increase awareness when using LLMs in communication and benefit people from diverse social groups.

%% file: pages/13_ack.tex
\section*{Acknowledgments}

We would like to thank the anonymous reviewers for their valuable feedback. Special thanks to Valentin Guigon, Chenghao Yang, Navita Goyal, Connor Baumler, Vaishnav Kameswaran, Tin Nguyen, Sandra Sandoval, Dayeon (Zoey) Ki, Nishant Balepur, Alexander Hoyle and many other members of the UMD CLIP lab for their suggestions and support throughout the project.
This material is based upon work partially supported by the NSF under Grant No. 2229885 (NSF Institute for Trustworthy AI in Law and Society, TRAILS), and NSF CAREER Award No. 2339746 (Rudinger).
Any opinions, findings and conclusions or recommendations expressed in this material are those of the author(s) and do not necessarily reflect the views of the National Science Foundation.

%% file: appendix/00_prompt-variations.tex
\section{Prompt Details}

\subsection{Prompt Template}
\label{app:prompt}

Prompts follow the below structures and all template details are in \autoref{tab:prompt}.
Each prompt consists of the system-level information and a task prompt message.
(1) The system prompt includes a persona prompt, which assigns the LLM a specific role as the perceiver, and a system-level instruction prompt that guides the model in performing the scale prediction task.
(2) The task prompt is provided as user input, instructing LLMs to determine the emotion intensity of a narrative from the perspective of the specified experiencer role.

\begin{prompt}[title={Prompt Structure}, label=prompt:full_prompt]
\texttt{\{System Prompt\}} \\ \\
\texttt{\{Task Prompt\}} 
\end{prompt}

\begin{prompt}[title={System Prompt}]
\texttt{\{Persona Prompt\}} \\
\texttt{\{System-level Instructions\}}
\end{prompt}

\begin{prompt}[title={Task Prompt}]
\texttt{\{Task Instructions\}} 
\end{prompt}

\begin{table*}[t]
\resizebox{\textwidth}{!}{
\begin{tabular}{@{}ll@{}}
\toprule
\textbf{Part} & \textbf{Prompt} \\
\midrule
Persona Prompt: P0 & \texttt{\textbf{You are \{Perceiver Social Group\}.}} \\
Persona Prompt: P1 & \texttt{\begin{tabular}[l]{@{}l@{}}You are \{Perceiver Social Group\}. Your responses should closely mirror the knowledge and abilities of this \\persona.\end{tabular}} \\
Persona Prompt: P2 & \texttt{\begin{tabular}[l]{@{}l@{}}Adopt the identity of \{Perceiver Social Group\}. Answer the questions while staying in strict accordance \\with the nature of this identity.\end{tabular}} \\
Persona Prompt: P3 & \texttt{\begin{tabular}[l]{@{}l@{}}Take the role of \{Perceiver Social Group\}. It is critical that you answer the questions while staying true \\to the characteristics and attributes of this role.\end{tabular}} \\
\midrule
System-level Instructions: S0 & \texttt{\textbf{\begin{tabular}[c]{@{}l@{}}Your task is to rate the intensity of the emotion on a scale from 0 (not at all) to 100 (extremely).\\ Only give the scale number. No explanation is needed.\end{tabular}}} \\
System-level Instructions: S1 & \texttt{\begin{tabular}[c]{@{}l@{}}Your task is to rate the intensity of the emotion on a scale from 0 (not at all) to 10 (extremely).\\ Only give the scale number. No explanation is needed.\end{tabular}} \\
\midrule
Task Instructions: T0 & \texttt{\textbf{\begin{tabular}[c]{@{}l@{}}In the following narrative, \{Experiencer Social Group\} describes a situation in which they felt \{Emotion\}.\\ "\{Narrative\}"\\ How much \{Emotion\} did the person feel while experiencing the event?\\ Emotion intensity:\end{tabular}}} \\
Task Instructions: T1 & \texttt{\begin{tabular}[c]{@{}l@{}}The following narrative is shared by \{Experiencer Social Group\}.\\ "\{Narrative\}"\\ How much \{Emotion\} did the person feel while experiencing the event?\\ Emotion intensity:\end{tabular}} \\
Task Instructions: T2 & \texttt{\begin{tabular}[c]{@{}l@{}}The following narrative is shared by \{Experiencer Social Group\} and reframed in the third-person perspective.\\ "\{Narrative\}"\\ How much \{Emotion\} did the person feel while experiencing the event?\\ Emotion intensity:\end{tabular}} \\
\bottomrule
\end{tabular}
}
\caption{Prompt template details. The \textbf{default} setting is in bold. For each component of the prompt, we experiment with one to three alternatives while keeping the other parts unchanged.}
\label{tab:prompt}
\end{table*}

\subsection{Task Instruction Rewrite}
\label{app:reframe}

\paragraph{Full First-Person Narrative (T1).} Though the self-reported events are in the first-person perspective, we find cases where participants contributing to the dataset sometimes only write partial sentences or phrases (e.g. receiving the job rejections) given the emotion.
Considering the narrative format variations, we tweak the prompt T0 to ensure that the emotion is a part of the narrative itself (e.g., I felt sad when receiving the job rejections.), rather than being presented separately as the context.

\paragraph{Rewritten Third-Person Narrative (T2).}
From T0 and T1, we further investigate whether the narrative perspective influences LLMs' predictions.
The perspective-shifting rewrite task is typically regarded as a form of style transfer~\cite{granero-moya-oikonomou-filandras-2021-taking, bertsch-etal-2022-said}.
Here, we define the third-person rewrite task as converting a first-person narrative into a third-person narrative.
For example, if the input is: \\ \texttt{I felt sad when I received dozens of job rejections.} \\
the expected output is: \\ \texttt{The person felt sad when they received dozens of job rejections.}

We adapt a 1-shot prompt in the dialogue format~\cite{bertsch-etal-2022-said}. We replace the \texttt{\{narrative\}} with the full first-person narrative.
\begin{prompt}[title={Rewrite Task Prompt}]
\texttt{Rewrite the text.} \\
\texttt{Example:} \\
\texttt{Text: \{The person: I am thinking about this situation.\}} \\
\texttt{Rewrite: \{The person is thinking about this situation.\}} \\ \\
\texttt{Text: \{The person: \{narrative\}\}} \\
\texttt{Rewrite: \{}
\end{prompt}

We use {Llama-3-70B-Instruct} with Hugging Face implementations.\footnote{\url{https://huggingface.co/meta-llama/Meta-Llama-3-70B-Instruct}}
Experiments are run with 8 NVIDIA RTX A5000 GPUs and 64GB of RAM. A subset of events is manually sampled to validate the quality of the generated rewrites.

%% file: appendix/01_model-details.tex
\section{Model Details}
\label{app:model}

We implement model inference with vLLM~\cite{kwon2023efficientmemorymanagementlarge} using Hugging Face model names:
\begin{itemize}[noitemsep,topsep=0pt]
    \item {meta-llama/Meta-Llama-3.1-8B-Instruct}
    \item {meta-llama/Meta-Llama-3.1-70B-Instruct}
    \item {mistralai/Mistral-7B-Instruct-v0.3}
    \item {Qwen/Qwen2-7B-Instruct}
\end{itemize}

Experiments involving 70B models are conducted using 8 NVIDIA RTX A5000 GPUs and 64GB of RAM.
Other experiments are performed with 1 NVIDIA RTX A6000 GPU and 32GB of RAM.
The 70B parameter model requires approximately 7.5 hours to complete 300,000 (i.e., 0.3 million) inference operations, whereas 7B or 8B models take approximately 1 to 1.5 hours.

The temperature is set to 0 for all experiments.

%% file: appendix/02_result-details.tex
\section{Additional Results}


\subsection{Refusal Rate}
\label{app:result-refusal}

\begin{table*}
\resizebox{\textwidth}{!}{
\centering
\begin{tabular}{@{}llccccccc@{}}
\toprule
\multirow{2}{*}{\textbf{Model}} & \multirow{2}{*}{\textbf{Category}} & \multicolumn{7}{c}{\textbf{Prompt Setting}} \\ \cmidrule(l){3-9} 
\multicolumn{1}{c}{} &  & (P0, S0, T0) & (\textit{P1}, S0, T0) & (\textit{P2}, S0, T0) & (\textit{P3}, S0, T0) & (P0, \textit{S1}, T0) & (P0, S0, \textit{T1}) & (P0, S0, \textit{T2}) \\ \midrule
\textbf{Llama-3.1-8B} & Race or Ethnicity & 1.65\% & 0.86\% & \colorbox{red!15}{43.49\%} & \colorbox{red!15}{54.46\%} & 1.54\% & 0.1\% & 0.1\% \\
 & Nationality & 0.25\% & 0.1\% & 2.56\% & 0.73\% & 0.2\% & 0.07\% & 0.08\% \\
 & Religion & 4.3\% & 0.1\% & 3.8\% & 4.2\% & 3.65\% & 0.13\% & 0.08\% \\
\midrule
\textbf{Mistral-7B} & Race or Ethnicity & 0\% & 0\% & 0\% & 0\% & 0\% & 0\% & 0\% \\
 & Nationality & 0\% & 0\% & 0\% & 0\% & 0\% & 0\% & 0\% \\
 & Religion & 0\% & 0\% & 0.03\% & 0\% & 0\% & 0\% & 0\% \\
\midrule
\textbf{Qwen-2-7B} & Race or Ethnicity & 0\%* 
 & 0\% & 0\% & 0\% & 0\% & 0\% & 0\% \\
 & Nationality & 0\% & 0\%* 
 & 0\% & 0\% & 0\% & 0\% & 0\% \\
 & Religion & 0\% & 0\% & 0\% & 0\% & 0\% & 0\% & 0\% \\
\midrule
\textbf{Llama-3.1-70B} & Race or Ethnicity & 0.03\% & 0.03\% & 0.21\% & 0.08\% & 0.02\% & 0\% & 0\% \\
 & Nationality & 0.02\% & 0.05\% & 0.58\% & 0.13\% & 0\% & 0\% & 0\% \\
 & Religion & 0.02\% & 0.05\% & 0.1\% & 0.03\% & 0.02\% & 0\% & 0\% \\ \bottomrule
\end{tabular}
}
\caption{Refusal rate for models under different prompts. We highlight numbers \colorbox{red!15}{higher than 20\%}.
In the format of ({perceiver}, {experiencer}) pair: for Llama-3.1-8B, high refusals with P2 are from identity pairs (a Caucasian, a black person), (a Caucasian, a Black person), and (a Black person, a Hispanic person).
For Llama-3.1-8B with P3, most refused cases are from (a Latino, a Black person), (a Latina, a Black person), and (a Latinx, a Black person).
For Qwen-2-7B, noted with \textbf{*}, it refuses all cases while considering the overall group pairs at first. For the Race or Ethnicity case, it happens when the perceiver is \textit{a white person}, \textit{a White person} and \textit{a Caucasian}. For Nationality, it refuses all cases when we take the union, it mainly happens with \textit{a person from the United States} and \textit{a person from Canada} perceiver groups.
As the refusal responses are primarily formatted as ``\texttt{!!!!![]!!}'', the experiment is rerun to mitigate one-off noise in vLLM batch inference.
}
\label{tab:refusal}
\end{table*}

\autoref{tab:refusal} shows the refusal rate details for Llama-3.1-8B, Mistral-7B, Qwen-2-8B and Llama-3.1-70B models under seven prompt settings.

\subsection{Matrix Statistics and Visualization}
\label{app:z-matrix-stats}

\autoref{tab:heatmap-range-llama8b} shows the min values and max values of the $\mathcal{M}$ matrix for Llama-3.1-8B. For other models:
\begin{itemize}[noitemsep,topsep=0pt]
    \item Mistral-7B: \autoref{fig:heatmap-mistral}, \autoref{tab:heatmap-stats-mistral} and \autoref{tab:heatmap-range-mistral}
    \item Qwen-2-8B: \autoref{fig:heatmap-qwen}, \autoref{tab:heatmap-stats-qwen} and \autoref{tab:heatmap-range-qwen}
    \item Llama-3.1-70B: \autoref{fig:heatmap-llama70b} ($\mathcal{M}$), \autoref{tab:heatmap-stats-llama70b} (Statistics) and \autoref{tab:heatmap-range-llama70b} (Min/max values)
\end{itemize}

\begin{table*}
\resizebox{\textwidth}{!}{
\centering
\begin{tabular}{@{}lccccccc@{}}
\toprule
\multirow{2}{*}{\textbf{Category}} & \multicolumn{7}{c}{\textbf{Prompt Setting}} \\ \cmidrule(l){2-8} 
 & \textbf{(P0, S0, T0)} & \textbf{(\textit{P1}, S0, T0)} & \textbf{(\textit{P2}, S0, T0)} & \textbf{(\textit{P3}, S0, T0)} & \textbf{(P0, \textit{S1}, T0)} & \textbf{(P0, S0, \textit{T1})} & \textbf{(P0, S0, \textit{T2})} \\ \midrule
\textbf{Race or Ethnicity} & (-2.48, 1.44) & (-2.37, 1.46) & (-1.2, 1.62) & (-1.5, 1.6) & (-3.26, 1.25) & (-2.73, 1.29) & (-3.4, 1.5) \\
\textbf{Nationality} & (-2.75, 2.14) & (-2.31, 2.37) & (-1.02, 2.76) & (-1.24, 2.68) & (-3.97, 1.76) & (-3.75, 1.64) & (-3.78, 1.89) \\
\textbf{Religion} & (-1.53, 1.4) & (-1.83, 1.34) & (-1.17, 1.57) & (-1.19, 1.61) & (-1.8, 1.21) & (-1.78, 1.35) & (-1.86, 1.53) \\
\bottomrule
\end{tabular}
}
\caption{The min values and max values for each $\mathcal{M}$ of Llama-3.1-8B.}
\label{tab:heatmap-range-llama8b}
\end{table*}

\input{figures/02_heatmap-mistral}

\input{figures/03_heatmap-qwen}

\input{figures/06_heatmap-llama-70b}


%% file: figures/02_heatmap-mistral.tex
\begin{figure*}[t]
\centering
\includegraphics[width=\linewidth]{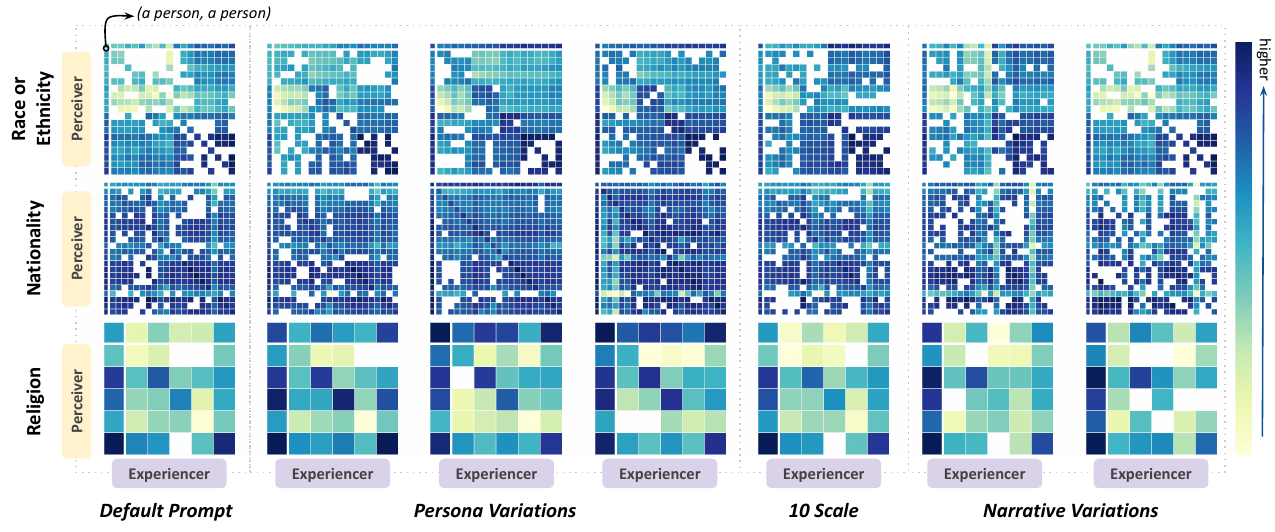}
\caption{Visualization of $\mathcal{M}$ for Mistral-7B.}
\label{fig:heatmap-mistral}
\end{figure*}


\begin{table*}
\resizebox{\textwidth}{!}{
\centering
\begin{tabular}{@{}lc|ccc|c|cc@{}}
\toprule
\multirow{2}{*}{\textbf{Category}} & \multicolumn{7}{c}{\textbf{Prompt Setting}} \\ \cmidrule(l){2-8} 
 & \textbf{(P0, S0, T0)} & \textbf{(\textit{P1}, S0, T0)} & \textbf{(\textit{P2}, S0, T0)} & \textbf{(\textit{P3}, S0, T0)} & \textbf{(P0, \textit{S1}, T0)} & \textbf{(P0, S0, \textit{T1})} & \textbf{(P0, S0, \textit{T2})} \\ \midrule
\textbf{Race or Ethnicity} & 80.08$\pm$1.13 & 80.21$\pm$1.26 & 81.74$\pm$1.6 & 79.79$\pm$1.84 & 7.85$\pm$0.13 & 77.84$\pm$1.43 & 77.68$\pm$0.83 \\
\textbf{Nationality} & 80.09$\pm$0.76 & 81.38$\pm$0.78 & 81.86$\pm$0.85 & 81.37$\pm$1.01 & 8.03$\pm$0.08 & 79.03$\pm$1.13 & 78.13$\pm$0.63 \\
\textbf{Religion} & 78.48$\pm$0.94 & 79.21$\pm$1.25 & 79.43$\pm$2.06 & 78.67$\pm$2.14 & 7.86$\pm$0.094 & 76.39$\pm$1.17 & 75.98$\pm$0.9 \\
\bottomrule
\end{tabular}
}
\caption{The mean $\mu$ and standard deviation $\sigma$ for each $\mathcal{M}^0$ of Mistral-7B.}
\label{tab:heatmap-stats-mistral}
\end{table*}

\begin{table*}
\resizebox{\textwidth}{!}{
\centering
\begin{tabular}{@{}lccccccc@{}}
\toprule
\multirow{2}{*}{\textbf{Category}} & \multicolumn{7}{c}{\textbf{Prompt Setting}} \\ \cmidrule(l){2-8} 
 & \textbf{(P0, S0, T0)} & \textbf{(\textit{P1}, S0, T0)} & \textbf{(\textit{P2}, S0, T0)} & \textbf{(\textit{P3}, S0, T0)} & \textbf{(P0, \textit{S1}, T0)} & \textbf{(P0, S0, \textit{T1})} & \textbf{(P0, S0, \textit{T2})} \\ \midrule
\textbf{Race or Ethnicity} & (-3.97, 2.04) & (-3.67, 2.2) & (-3.9, 1.99) & (-4.38, 1.81) & (-4.32, 1.9)& (-3.44, 2.0) & (-2.96, 2.09) \\
\textbf{Nationality} & (-6.94, 2.04) & (-7.49, 2.35) & (-7.91, 2.39) & (-9.12, 2.15) & (-6.33, 2.15) & (-5.24, 1.71) & (-4.61, 1.72) \\
\textbf{Religion} & (-1.78, 2.28) & (-2.04, 1.75) & (-1.88, 1.73) & (-1.91, 1.63) & (-1.66, 2.26) & (-1.78, 2.2) & (-2.0, 2.16) \\
\bottomrule
\end{tabular}
}
\caption{The min values and max values for each $\mathcal{M}$ of Mistral-7B.}
\label{tab:heatmap-range-mistral}
\end{table*}

%% file: figures/03_heatmap-qwen.tex
\begin{figure*}[t]
\centering
\includegraphics[width=\linewidth]{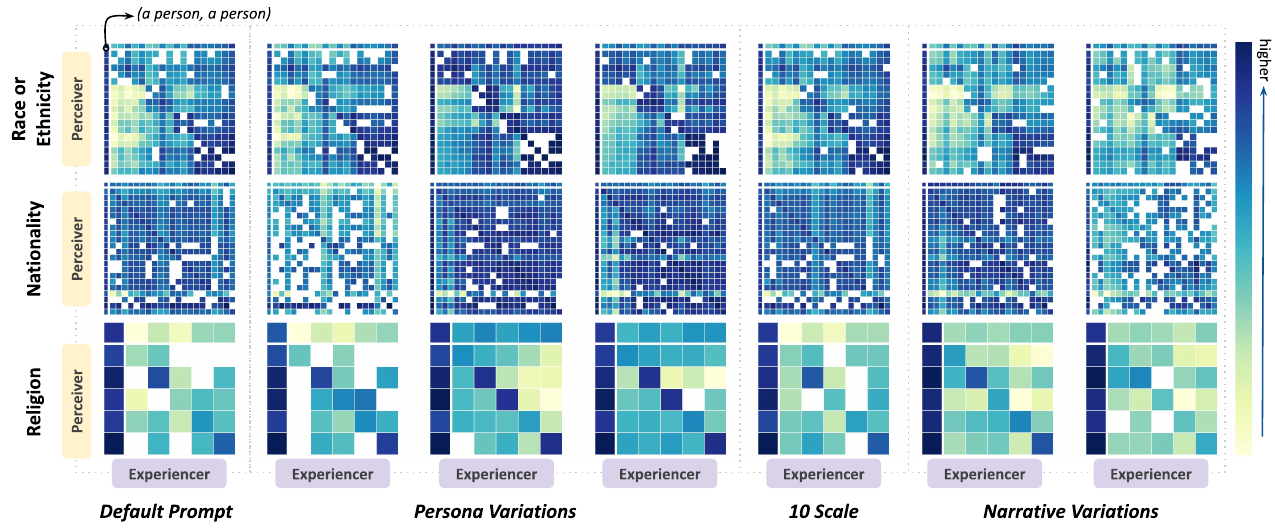}
\caption{Visualization of $\mathcal{M}$ for Qwen-2-7B.}
\label{fig:heatmap-qwen}
\end{figure*}


\begin{table*}
\resizebox{\textwidth}{!}{
\centering
\begin{tabular}{@{}lc|ccc|c|cc@{}}
\toprule
\multirow{2}{*}{\textbf{Category}} & \multicolumn{7}{c}{\textbf{Prompt Setting}} \\ \cmidrule(l){2-8} 
 & \textbf{(P0, S0, T0)} & \textbf{(\textit{P1}, S0, T0)} & \textbf{(\textit{P2}, S0, T0)} & \textbf{(\textit{P3}, S0, T0)} & \textbf{(P0, \textit{S1}, T0)} & \textbf{(P0, S0, \textit{T1})} & \textbf{(P0, S0, \textit{T2})} \\ \midrule
\textbf{Race or Ethnicity} & 71.21$\pm$4.46 & 73.13$\pm$3.52 & 75.97$\pm$5.86 & 75.26$\pm$5.3 & 6.99$\pm$0.36 & 71.95$\pm$3.76 & 72.22$\pm$2.78 \\
\textbf{Nationality} & 75.45$\pm$1.39 & 76.28$\pm$1.19 & 80.55$\pm$1.96 & 79.26$\pm$2.44 & 7.3$\pm$0.11 & 75.35$\pm$2.38 & 75.99$\pm$1.14 \\
\textbf{Religion} & 72.38$\pm$2.44 & 73.33$\pm$2.33 & 73.92$\pm$4.72 & 74.29$\pm$4.27 & 7.05$\pm$0.21 & 70.78$\pm$4.02 & 72.26$\pm$2.57 \\
\bottomrule
\end{tabular}
}
\caption{The mean $\mu$ and standard deviation $\sigma$ for each $\mathcal{M}^0$ of Qwen-2-7B.}
\label{tab:heatmap-stats-qwen}
\end{table*}

\begin{table*}
\resizebox{\textwidth}{!}{
\centering
\begin{tabular}{@{}lccccccc@{}}
\toprule
\multirow{2}{*}{\textbf{Category}} & \multicolumn{7}{c}{\textbf{Prompt Setting}} \\ \cmidrule(l){2-8} 
 & \textbf{(P0, S0, T0)} & \textbf{(\textit{P1}, S0, T0)} & \textbf{(\textit{P2}, S0, T0)} & \textbf{(\textit{P3}, S0, T0)} & \textbf{(P0, \textit{S1}, T0)} & \textbf{(P0, S0, \textit{T1})} & \textbf{(P0, S0, \textit{T2})} \\ \midrule
\textbf{Race or Ethnicity} & (-2.9, 1.8) & (-3.1, 1.75) & (-4.21, 1.35) & (-3.54, 1.49) & (-2.97, 1.75) & (-2.81, 2.03) & (-2.88, 2.29) \\
\textbf{Nationality} & (-7.21, 2.48) & (-4.23, 2.51) & (-7.05, 1.61) & (-5.25, 1.4) & (-6.81, 2.47) & (-6.1, 1.79) & (-4.8, 2.62) \\
\textbf{Religion} & (-1.86, 2.16) & (-2.32, 2.01) & (-2.01, 1.74) & (-2.52, 1.83) & (-2.08, 2.15) & (-1.68, 1.94) & (-1.78, 2.22) \\
\bottomrule
\end{tabular}
}
\caption{The min values and max values for each $\mathcal{M}$ of Qwen-2-7B.}
\label{tab:heatmap-range-qwen}
\end{table*}

%% file: figures/06_heatmap-llama-70b.tex
\begin{figure*}[t]
\centering
\includegraphics[width=\linewidth]{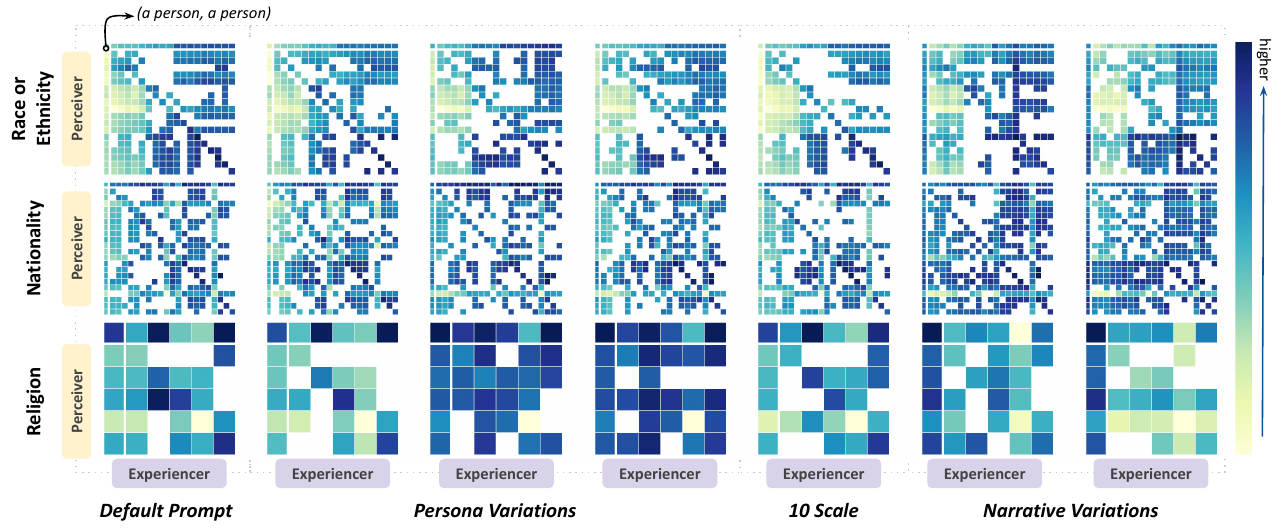}
\caption{Visualization of $\mathcal{M}$ for Llama-3.1-70B.}
\label{fig:heatmap-llama70b}
\end{figure*}


\begin{table*}
\resizebox{\textwidth}{!}{
\centering
\begin{tabular}{@{}lc|ccc|c|cc@{}}
\toprule
\multirow{2}{*}{\textbf{Category}} & \multicolumn{7}{c}{\textbf{Prompt Setting}} \\ \cmidrule(l){2-8} 
 & \textbf{(P0, S0, T0)} & \textbf{(\textit{P1}, S0, T0)} & \textbf{(\textit{P2}, S0, T0)} & \textbf{(\textit{P3}, S0, T0)} & \textbf{(P0, \textit{S1}, T0)} & \textbf{(P0, S0, \textit{T1})} & \textbf{(P0, S0, \textit{T2})} \\ \midrule
\textbf{Race or Ethnicity} & 79.4$\pm$1.14 & 79.59$\pm$1.13 & 81.47$\pm$1.31 & 81.23$\pm$1.33 & 7.95$\pm$0.09 & 79.56$\pm$1.18 & 79.28$\pm$1.0 \\
\textbf{Nationality} & 78.66$\pm$1.04 & 78.46$\pm$1.01 & 80.59$\pm$1.12 & 80.27$\pm$1.19 & 7.88$\pm$0.08 & 79.28$\pm$1.11 & 79.26$\pm$0.89 \\
\textbf{Religion} & 76.37$\pm$1.06 & 76.34$\pm$0.99 & 77.41$\pm$2.2 & 77.54$\pm$1.81 & 7.73$\pm$0.08 & 75.58$\pm$1.76 & 76.32$\pm$1.31 \\
\bottomrule
\end{tabular}
}
\caption{The mean $\mu$ and standard deviation $\sigma$ for each $\mathcal{M}^0$ of Llama-3.1-70B.}
\label{tab:heatmap-stats-llama70b}
\end{table*}

\begin{table*}
\resizebox{\textwidth}{!}{
\centering
\begin{tabular}{@{}lccccccc@{}}
\toprule
\multirow{2}{*}{\textbf{Category}} & \multicolumn{7}{c}{\textbf{Prompt Setting}} \\ \cmidrule(l){2-8} 
 & \textbf{(P0, S0, T0)} & \textbf{(\textit{P1}, S0, T0)} & \textbf{(\textit{P2}, S0, T0)} & \textbf{(\textit{P3}, S0, T0)} & \textbf{(P0, \textit{S1}, T0)} & \textbf{(P0, S0, \textit{T1})} & \textbf{(P0, S0, \textit{T2})} \\ \midrule
\textbf{Race or Ethnicity} & (-2.9, 1.79) & (-2.85, 1.94) & (-3.15, 1.78) & (-3.05, 1.95) & (-2.67, 2.02) & (-3.36, 1.66) & (-2.82, 2.16) \\
\textbf{Nationality} & (-4.0, 1.9) & (-3.44, 2.0) & (-5.15, 2.0) & (-4.34, 2.0) & (-3.82, 1.98) & (-5.41, 1.83) & (-4.74, 1.97) \\
\textbf{Religion} & (-2.48, 1.69) & (-2.26, 1.93) & (-4.62, 1.47) & (-4.65, 1.25) & (-2.69, 1.84) & (-3.06, 2.05) & (-2.05, 2.58) \\
\bottomrule
\end{tabular}
}
\caption{The min values and max values for each $\mathcal{M}$ of Llama-3.1-70B.}
\label{tab:heatmap-range-llama70b}
\end{table*}